\documentclass[runningheads]{llncs}
\usepackage{multirow}
\usepackage{graphicx}
\usepackage{booktabs}
\usepackage{xr}

\usepackage{eccv}

\usepackage{eccvabbrv}

\usepackage{graphicx}
\usepackage{booktabs}
\usepackage{multicol}
\usepackage{multirow}
\usepackage{tabularx}
\usepackage{siunitx} 
\usepackage{footnote}

\usepackage{array} 

\newcolumntype{C}{>{\centering\arraybackslash}X}

\usepackage[accsupp]{axessibility}  

\usepackage{hyperref}

\usepackage{orcidlink}

\usepackage{soul}
\usepackage{enumitem}
\usepackage{array}

\begin{document}
\usetikzlibrary{positioning, shapes, arrows.meta, shadows, shapes.geometric, calc, arrows.meta, shapes.misc}
\title{Diffusion Soup: Model Merging for Text-to-Image Diffusion Models} 


\author{Benjamin Biggs$^*$\inst{1} \and
Arjun Seshadri$^*$\inst{1} \and
Yang Zou\inst{2} \and
Achin Jain\inst{2} \and
Aditya Golatkar\inst{1} \and
Yusheng Xie\inst{2} \and
Alessandro Achille\inst{1} \and
Ashwin Swaminathan\inst{2} \and
Stefano Soatto\inst{1}}

\authorrunning{B.~Biggs et al.}

\institute{AWS AI Labs \and
Amazon AGI Foundations}

\maketitle
\def\thefootnote{*}\footnotetext{Equal contribution, alphabetical order. Correspondence to: \texttt{benbiggs@amazon.com.}}\def\thefootnote{\arabic{footnote}}

\begin{abstract}
We present Diffusion Soup, a compartmentalization method for Text-to-Image Generation that averages the weights of diffusion models trained on sharded data. By construction, our approach enables training-free continual learning and unlearning with no additional memory or inference costs, since models corresponding to data shards can be added or removed by re-averaging. We show that Diffusion Soup samples from a point in weight space that approximates the geometric mean of the distributions of constituent datasets, which offers anti-memorization guarantees and enables zero-shot style mixing. Empirically, Diffusion Soup outperforms a paragon model trained on the union of all data shards and achieves a 30\% improvement in Image Reward (.34 $\to$ .44) on domain sharded data, and a  59\% improvement in IR (.37 $\to$ .59) on aesthetic data. In both cases, souping also prevails in TIFA score (respectively, 85.5 $\to$ 86.5 and 85.6 $\to$ 86.8). We demonstrate robust unlearning---removing any individual domain shard only lowers performance by 1\% in IR (.45 $\to$ .44)---and validate our theoretical insights on anti-memorization using real data. Finally, we showcase Diffusion Soup's ability to blend the distinct styles of models finetuned on different shards, resulting in the zero-shot generation of hybrid styles.
\end{abstract}

\section{Introduction}\label{sec:intro}

On paper, the ultimate goal of a foundational image generation model is to approximate, given a large dataset $\mathcal{D}$, the underlying data distribution $p(x)$ generating those images. This view, however, hides the important subtleties that come with real-world data: any real large scale training dataset $\mathcal{D}$ is not just an identically distributed collection of images, but rather a variable entity where new data, coming from different sources covering different domains and usage rights, is frequently added or removed. In these situations, training a monolithic model on all data is problematic. If the data changes, the whole model has to be retrained (at a large cost) to either add new information (continual learning), or remove information that cannot be used anymore (machine unlearning). Moreover, while a single model trained on all the data together can have impressive coverage and overall performance, it often underperforms expert models trained on particular domains (see Table \ref{table:mixture_10k_new}), to the detriment of downstream users interested in those particular use-cases.

\begin{figure*}[t] 
\centering
\includegraphics[trim=60 10 110 0, clip, width=\linewidth]{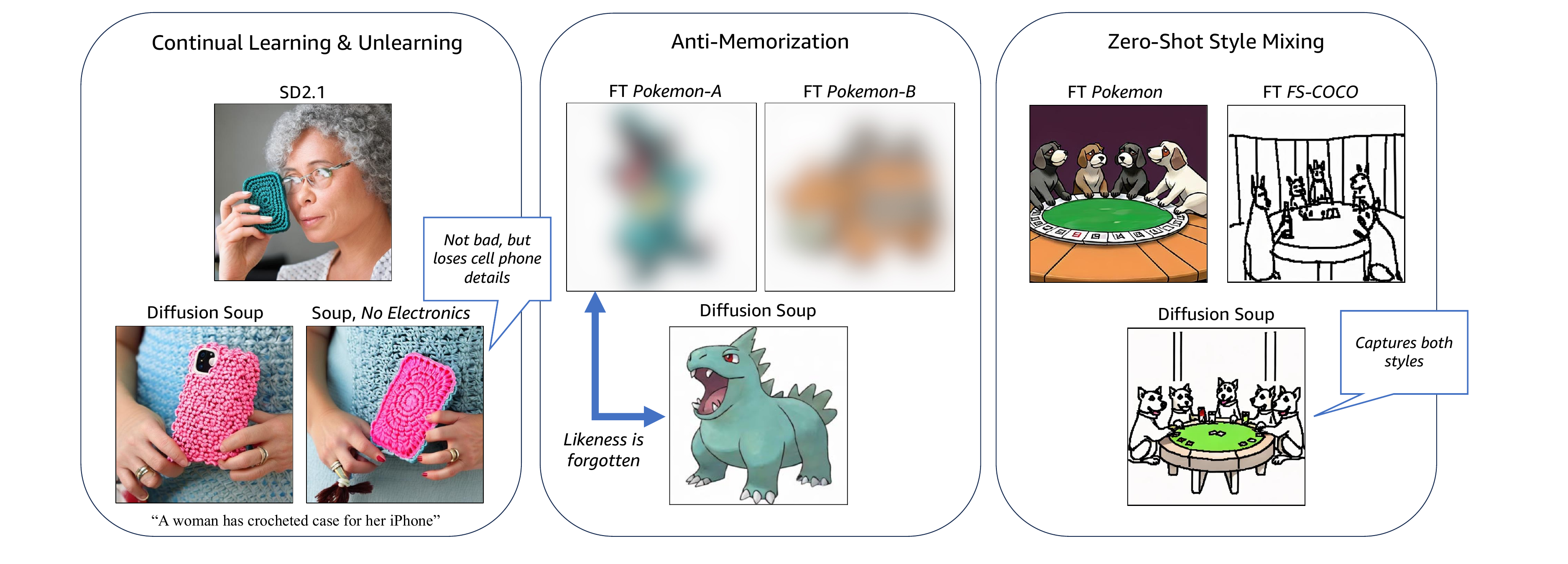}
\caption{\textbf{Diffusion Soup Enables Three Distinct Applications.} (1) \emph{Continual Learning \& Unlearning}: models trained on various data shards can be added to improve performance or subtracted when removal is necessary. (2) \emph{Zero-Shot Style Mixing}: souping blends the styles into a hybrid of its components with no extra supervision. (3) \emph{Anti-Memorization}: Diffusion Soup prevents memorization while capturing its high level style (note that we blur depictions of inputs in this subfigure).}
\label{fig:splash}
\end{figure*}

\begin{figure*}[t] 
\centering
\includegraphics[trim=0 0 200 0, clip, width=\linewidth]
{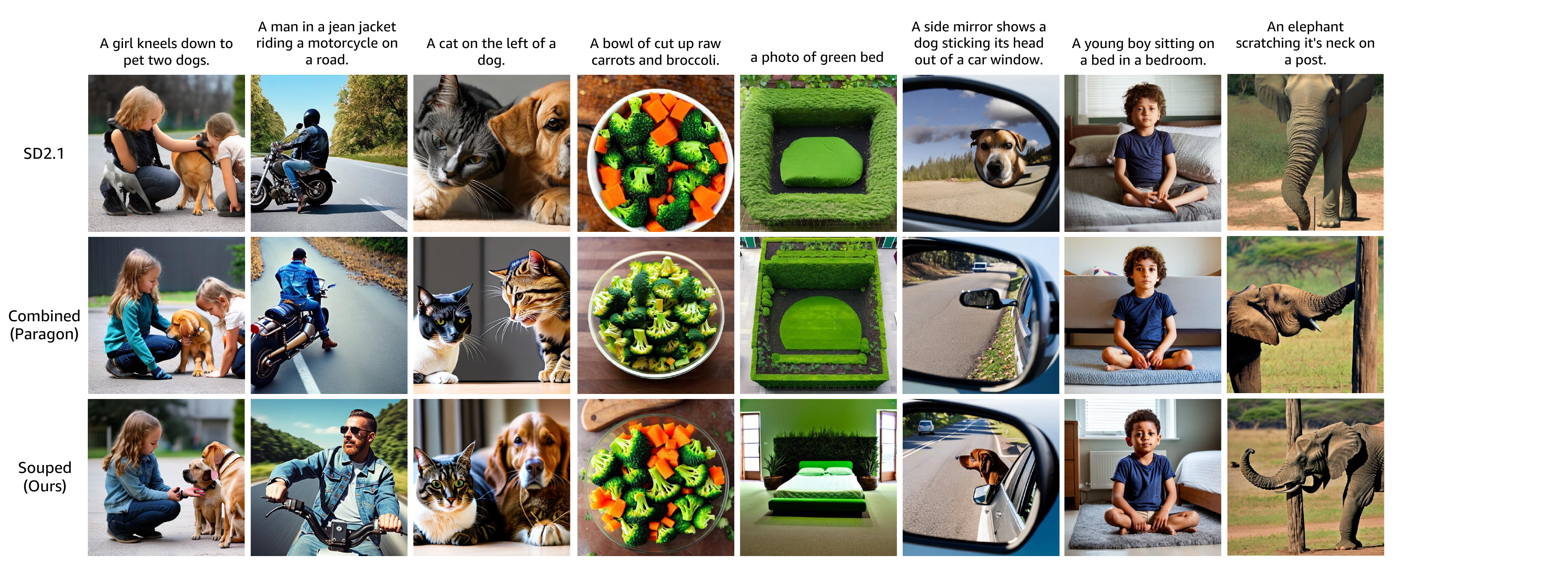}
\caption{\textbf{Images from Diffusion Soup Beat SD2.1 and a Combined Paragon.} We visualize images generated by averaging the weights of various models finetuned on data shards spanning different categories (\emph{Souped}), and compare them to images from the pretrained model (\emph{SD2.1}) and a paragon model trained all the shards (\emph{Combined}). These images highlight Diffusion Soup's dominance in metrics (See Table \ref{table:mixture_10k_new}) for Text-Image Alignment, Aesthetics, and Fidelity. Best viewed in color and zoomed in.}
\label{fig:soup_vs_others}
\end{figure*}

This dilemma has prompted interest in \textit{compartmentalization} or \textit{mixture of expert} (MoE) methods: rather than treating the model as a monolithic blackbox trained on all data, different sets of parameters are trained on different shards of the training set (corresponding, for example, to different domains or data provided by different users) and then combined at inference time. Compartmentalized models excel at continual learning, unlearning, and anti-memorization tasks. However, in the simplest implementation through ensembling of independent models, they significantly increase inference time --- especially as the number of models grows in the hundreds/thousands as would be desirable to have fine control over data provenance. Moreover, ensemble models are significantly more complex to deploy, as they require custom architectures and proper load distribution across instances. The natural question is whether there is a simple method to merge information between different models trained on different subsets of data into a single model, with little or no additional computational overhead.

To answer this question, we introduce \textit{Diffusion Soup}. Diffusion Soup trains separate models on different subsets of data, and then simply averages their weights to obtain a single model. Averaging weights may seem ill-advised, since weights do not live in a linear (vector) space and therefore averaged weights may be far from, and perform worse than, any of its components. However, we show that, when done properly, this strategy not only leads to viable models, but actually outperforms a paragon monolithic model trained on the combination of all data. See \cref{fig:soup_vs_others} for a visualization of results from Diffusion Soup alongside a base model and the paragon.

Our result leverages the insight that the training process {\em enforces} linearity by means of optimization with gradient descent: Each weight along the training path is modified by an {\em additive} (linear) increment in a randomized direction. Exploiting the structure of the update, we can show via Taylor expansion that a Diffusion Soup of models finetuned from a shared pretrained checkpoint is expected to (approximately) sample images from the geometric mean of the individual model distributions. In contrast, training a monolithic model samples from their arithmetic mean. Sampling from the geometric mean acts as an implicit regularizer, which counters the decrease in quality due to the overfitting commonly observed during fine-tuning of diffusion models. This justifies our counter-intuitive results that a model soup performs better than a monolithic expert trained with access to all the data.\footnote{While our results are valid for local perturbations around a pretrained point, empirically we show that the common pre-trained point can be generic enough that individually trained experts starting from it achieve comparable performance than if they were trained from scratch.}

Empirically, we find that Diffusion Soup outperforms the comparable paragon model trained on all the data shards in two settings: (1) when data shards are grouped by domain (e.g. \emph{Animals}, \emph{Electronics}, etc.), Diffusion Soup excels in TIFA score (85.5 $\to$ 86.5) and Image Reward (0.34 $\to$ 0.44); and (2) when data shards are all geared towards Aesthetics, Diffusion Soup prevails in TIFA score (85.6 $\to$ 86.8), Image Reward (0.37 $\to$ 0.59), and CLIP Score (0.261 $\to$ 0.263). 

We also show that Diffusion Soup: (1) can be used for training-free unlearning (simply removing weights from the average); removing \textit{any} domain's shard decreases performance by at most 1\% in Image Reward (.45 $\to$ .44); (2) the souped model approximates a Near Access Freeness condition, which provides the souped model with better anti-memorization guarantees; (3) can be modified by the user to achieve desired zero-shot \textit{blending} between styles. 
\section{Related Work}
\textit{Diffusion models.} 
Diffusion models \cite{ho2020denoising,dhariwal2021diffusion,salimans2022progressive,song2020denoising,ramesh2021zero} are state-of-the-art models used for text-to-image generation. These models add Gaussian noise to an image (or its latents) in the forward process, and learn the score function to denoise the image in the backward process (introduced by \cite{lindquist1979stochastic}) during generation. \cite{rombach2022high,ramesh2021zero} condition diffusion models on text to enable text-to-image generation at inference using a technique called classifier-free guidance\cite{ho2022classifier}. \cite{song2020score,song2021maximum} show that the forward-backward process obeys a stochastic differential equation, we use this framework in this paper.
\textit{Compositional generation.} 
\cite{liu2022compositional,du2023reduce,golatkar2023training,balaji2022ediffi,chen2022towards,zhou2022mixture} propose performing compositional generation by merging output flows of diffusion models, methods that often incur increased compute cost. \cite{xue2024raphael} proposes a mixture of expert method by replacing the standard convolutional layers with mixture-of-expert layers which reduce the inference cost and the number of parameters during inference. We show that our method further reduces the inference cost and parameters--and in fact incurs no additional inference or memory overhead beyond that of a single model. 
\textit{Model merging.} Weight averaging \cite{garipov2018loss,izmailov2018averaging,choshen2022fusing,matena2111merging,wortsman2022model} is a popular technique used to improve the accuracy of discriminative models. \cite{liu2023tangent,liu2023tangenttransformer} further extended these methods for linearized convolutional and transformer based models. \cite{avrahami2022gan} proposed a model averaging method for generative adversarial networks, however, such methods have been unexplored for diffusion models. To the best of our knowledge, we are the first method to show that weight averaging not only generates high quality images, but also improves the text-to-image alignment and reduces memorization at inference.
\label{sec:formatting}

\section{Preliminaries}
Latent diffusion models like Stable Diffusion~\cite{rombach2022high} aim to model a data distribution $p(x_0)$ of images $x_0 \in \mathcal{X}$ which can be sampled at inference time. We use the stochastic differential equations (SDE) based diffusion formalism of \cite{song2020score,karras2022elucidating} to define the basics of diffusion models. Given the initial distribution $p(x_0)$ at time $t=0$, SDE based formalism transforms it into a reference distribution $p(x_1) = \mathcal{N}(0,I)$ (Gaussian distribution) at time $t=1$ in the forward process:
\begin{equation}
dx_t = -\dfrac{1}{2}\beta_tx_tdt + \sqrt{\beta_t}d\rho_t
\label{eq:forward-diffusion}
\end{equation}
where $x_t$ is the diffused latents at $t$, $d\rho_t$ is the standard Wiener process, and $\beta_t$ are time varying diffusion coefficients used to manipulate the signal-to-noise ratio of the diffusion process. The intermediate latents $x_t$ are distributed according to a conditional gaussian distribution $p(x_t|x_0) = \mathcal{N}(x_t; \gamma_t x_0, \sigma_t^2I)$ by construction where $\gamma_t = \exp(-\dfrac{1}{2}\int_0^t\beta_tdt)$ and $\sigma_t^2 = 1 - \gamma_t^2$, providing the following marginal: $p(x_t) = \int_{x_0} p(x_t|x_0) dx_0$. \cite{lindquist1979stochastic,song2020score} shows that the forward process in \cref{eq:forward-diffusion} can be effectively reversed by the backward diffusion process given by:
\begin{equation}
dx_t = -\dfrac{1}{2}\beta_tx_tdt + \sqrt{\beta_t}d\rho_t - \nabla_{x_t}\log p(x_t)dt
\label{eq:backward-diffusion}
\end{equation}
where $dt$ is a decrement in time corresponding to the backward process, introduced by \cite{lindquist1979stochastic}. \cref{eq:backward-diffusion} reduces sampling from $p(x_0)$ to iterative application of $\nabla_{x_t}\log p(x_t)$ (score function) given an initial noisy samples $x_1 \sim \mathcal{N}(0,I)$. The score function $\nabla_{x_t}\log p(x_t)$ is generally difficult to estimate in practice and is instead learned using a neural network $\epsilon_{w}(x_t, t)$ and a training dataset $D$ through score-matching \cite{hyvarinen2005estimation,song2021maximum,dockhorn2021score,song2020score}. To enhance the usability and control of diffusion models they are often conditioned with textual prompts $y$ to model the conditional distribution $p(x_0|y)$ which modifies the score-estimating neural network as $\epsilon_w(x_t, t, y)$, to now accept $y$ as input during training and inference. Incorporating all these subtleties results in the following optimization problem to train text-to-image diffusion models:
\begin{equation}
\text{min}_w \mathbb{E}_{(x_0,y) \sim p(x_0,y)}\mathbb{E}_t [\|\epsilon_w(x_t, t, y) + \nabla_{x_t}\log p(x_t|x_0)\|]
\label{eq:score-matching}
\end{equation}

We can obtain an equivalent optimization problem by formulating the forward process in \cref{eq:forward-diffusion} as a Markov chain and optimizing a variant of the evidence lower bound (ELBO) \cite{ho2020denoising,song2020denoising}.
\section{Diffusion Soup}
While Diffusion models are pre-trained on large monolithic datasets (e.g., LAION-5B), in downstream applications it is often more common to have datasets $D_i$ come from different data sources, which may correspond to different data providers, or different domains. We use $\{p^{(i)}(x_0)\}_{i}$ to represent the collection of distributions corresponding to datasets, $\{D_i\}_{i=1}$.
Ensembling the outputs of generative models \cite{golatkar2023training,balaji2022ediffi} integrates information from different data sources, but is expensive due to the size of the models. Instead we propose to ensemble the weights of the model inspired by recent work \cite{lee2019wide,achille2021lqf,malladi2023kernel,golatkar2021mixed,wei2023ntk,zancato2020predicting,liu2023tangent} showing that the fine-tuning dynamics of the large models can be approximated with a Taylor series approximation: 
\begin{equation}
\epsilon_{w_c + \Delta w_i}(x_t, t, y) \approx \epsilon_{w_c}(x_t, t, y) + \nabla_{w} \epsilon_{w_c}(x_t, t, y)|_{w=w_c} \Delta w_i
\label{eq:linearization}
\end{equation}
where $\Delta w_i$ is the perturbation of the weights learned during fine-tuning. This result can be leveraged to show that ensembling the outputs corresponds to ensembling the weights of the models $w_i$ trained on different data sources $D_i$ (sampled from $p^{(i)}(x_0)$). More precisely, we define the souped prediction as:
\begin{align}
\epsilon_{\text{soup}} &\triangleq \underbrace{\epsilon_{\sum_i k_i w_i}}_{\text{Souping}}(x_t, t, y) = \epsilon_{\sum_i k_i (w_c + \Delta w_i)}(x_t, t, y) \nonumber \\
&\stackrel{(a)}{\approx} \epsilon_{w_c}(x_t, t, y) + \nabla_{w} \epsilon_{w_c}(x_t, t, y)|_{w=w_c} \cdot \Big(\sum_i k_i \Delta w_i\Big) \nonumber \\
&= \sum_i k_i \Big(\epsilon_{w_c}(x_t, t, y) + \nabla_{w} \epsilon_{w_c}(x_t, t, y)|_{w=w_c} \Delta w_i\Big) \nonumber \\
&\stackrel{(b)}{\approx}\sum_i k_i \epsilon_{w_c + \Delta w_i} = \sum_i k_i \epsilon_{w_i}(x_t, t, y) = \epsilon_{\text{ensemble}}\label{eq:model-soup}
\end{align}
where $k_i>0$, such that $\sum_i k_i = 1$ is a hyper-parameter which can be tuned by the user, and $(a),(b)$ follow from using the first order Taylor series approximation of the network. To summarize, Diffusion Soup approximates ensembling, and involves fine-tuning $n-$diffusion models ($\{\epsilon_{w_i}(x_t, t, y)\}_i$) on $n-$data sources ($\{p^{(i)}(x_0)\}_{i}$) respectively, and averaging the parameters.

\subsection{Sampling Distribution for Souping}

Modifying the score (or the $\epsilon(x_t)$) during the backward diffusion process changes the distribution which is sampled by the model. For instance, using \cref{eq:backward-diffusion} samples images from the distribution $p(x_0)$. In our case, since we modify the individual $\{\epsilon_{w_i}(x_t, t, y)\}$ to the soup $\epsilon_{\sum_i k_i w_i}(x_t, t, y)$ it is essential to identify the sampling distribution of the model. Depending on the choice of $k_i$ we can show that the souped model can either sample from various aggregations of the set of distributions $\{p^{(i)}(x_0)\}$. This is described in the following results:

\begin{proposition}(Geometric Mean)\label{prop:geo_mean}
Let $\nabla_{x_t} \log p^{(i)}(x_t)$ be the marginal score in \cref{eq:backward-diffusion} where  $x_t = \gamma_t x_0 + \sigma_t \epsilon$, with $x_0 \sim p^{(i)}(x_0)$ and $\epsilon \sim \mathcal{N}(0, I)$. Let $\epsilon_{w_i}(x_t, t, y)$ be a neural network with sufficient capacity trained to match $\nabla_{x_t} \log p^{(i)}(x_t)$. Under \cref{eq:backward-diffusion}'s conditions on the sampling procedure, $(1/n)\sum_{i=1}^n\nabla_{x_t} \log p^{(i)}(x_t)$ generates samples from $Z^{-1}\big(\prod_i p^{(i)}(x_0)\big)^{1/n}$. Furthermore, using \cref{eq:linearization}, $\epsilon_{\text{soup}}$ with $k_i = 1/n$ generates samples from $Z^{-1}\big(\prod_i p^{(i)}(x_0)\big)^{1/n}$.
\label{prop:geometric}
\end{proposition}
The previous result states Diffusion Soup with $k_i = 1/n$ generates samples from the geometric mean of the distribution of the individual models, which can be useful to provide memorization-free generation \cite{vyas2023provable}. Conversely, the following proposition shows souping (with an appropriately chosen value of $k_i$) samples from the union of all data (equivalent to training a model on the union).
\begin{proposition}(Arithmetic Mean)
Let $\nabla_{x_t} \log p^{(i)}(x_t)$ be the marginal score in \cref{eq:backward-diffusion} where  $x_t = \gamma_t x_0 + \sigma_t \epsilon$, with $x_0 \sim p^{(i)}(x_0)$ and $\epsilon \sim \mathcal{N}(0, I)$. Let $\epsilon_{w_i}(x_t, t, y)$ be a neural network (with sufficient capacity) trained to match $\nabla_{x_t} \log p^{(i)}(x_t)$. Sampling using $(1/n)\sum_{i=1}^n \lambda_i \frac{p^{(i)}(x_t)}{p(x_t)}\nabla_{x_t} \log p^{(i)}(x_t)$ with \cref{eq:backward-diffusion} generates samples from $\sum_i \lambda_i p^{(i)}(x_0)$, where $p(x_t) = \sum_i \lambda_i p^{(i)}(x_t)$, $\sum_i \lambda_i = 1$. Furthermore, using \cref{eq:linearization}, $\epsilon_{\text{soup}}$ with $k_i = \lambda_i \frac{p^{(i)}(x_t)}{p(x_t)}$ generates samples from $\sum_i \lambda_i p^{(i)}(x_0)$.
\label{prop:arithmetic}
\end{proposition}
Depending on the choice of $k_i$ in souping, we show that our model can sample from different distributions. Note that in \cref{prop:arithmetic}, $k_i$ depends on the $t$ and thus requires re-souping per prompt at every timestep. In this work, $k_i$ is fixed across prompts and timesteps, and set to be uniform ($k_i = 1/n$) or chosen using various greedy approaches, as we detail below.

\subsection{Greedy Souping}\label{sec:greedysoup}
In \cref{eq:model-soup} we show that the optimal weights after souping are a linear combination of the individual weights $w_i$ (data sources/experts), $w_{\text{soup}} = \sum_i k_i w_i$, where we treat $k_i$ as hyper-parameters to be optimized before inference subject to the constraint $\sum_i k_i = 1$. Let $L(w_{\text{soup}}, D_{\text{val}})$ be a evaluation metric used assess the quality of the diffusion model, for example, TIFA score or Image Reward. The following optimization problem solves for the optimal coefficients $k_i$:
\begin{equation}
\{k_i^*\}_i = \operatornamewithlimits{argmin}_{\sum_i k_i = 1, k_i > 0} \textstyle L\big(\sum_i k_i w_i, D_{\text{val}}\big)
\label{eq:coeff-opt-prob}
\end{equation}
Obtaining a closed form expression for \cref{eq:coeff-opt-prob} is intractable for diffusion model evaluation, and so we consider two greedy approaches to obtain coefficients. The first, Greedy Soup begins with the best performing individual weight and incrementally soups weights in order of individual performance---keeping the weights if they improve performance and discarding them if they do not. The second, Reverse Greedy Soup begins with the uniform soup and removes weights in order of increasing performance.

\section{Analysis}
\subsection{Near Access Freeness of Diffusion Soup}
Diffusion models often memorize some of their training data \cite{carlini2023extracting,somepalli2024understanding} posing a risk of reproducing samples from the training dataset. \cite{vyas2023provable} recently proposed a mathematical framework based on near-access freeness (NAF) to prevent generative models from memorizing samples present in the training data.
Let $D$ be the dataset with some samples $C=\{c_i\}_i$, and $D=D_1 \bigcup D_2$ be two disjoint splits of the dataset. Let $p^{(1)}(x|y), p^{(2)}(x|y)$ (where $x$ is the image and $y$ is the caption) be two diffusion models trained on $D_1, D_2$ respectively. Then Algorithm 3 from \cite{vyas2023provable} (CP$-\Delta$) shows that sampling from the geometric mean $p(x|y) = Z^{-1}\sqrt{p^{(1)}(x|y)p^{(2)}(x|y)}$ provides anti-memorization guarantees with respect to any sample $c_i$ present in only one of $D_1$ or $D_2$. That is, it satisfies $\varepsilon$-NAF:
\begin{equation}
\Delta(p(x|y)||\text{safe}_C(x|y)) \leq \varepsilon
\label{eq:cp-delta}
\end{equation}
where $\text{safe}_{C}(x|y)$ is a model which has not been trained on $c_i$, $\Delta$ is a divergence between two distributions (e.g., Kullback-Leibler divergence). Sampling from $p(x|y) = Z^{-1}\sqrt{p^{(1)}(x|y)p^{(2)}(x|y)}$ is difficult for diffusion models. However, sampling can be performed implicitly in the backward step during score computation \cite{golatkar2023training,du2023reduce} as $\nabla \log p(x|y) = (1/2)\big(\nabla \log p^{(1)}(x|y) + \nabla \log p^{(2)}(x|y)\big)$. With this observation we have the following result:
\begin{proposition}($\varepsilon$-NAF for Diffusion\ Soup)
Let $\epsilon_{\text{soup}}(x_t, t, y) \triangleq \epsilon_{\frac{1}{2}(w_1+w_2)}(x_t, t, y)$ be the soup model obtained by training $w_i$ on $D_i$, such that $D = D_1 \cup D_2$, and $D_1 \cap D_2 = \varnothing$. Then, under certain conditions, sampling using $\epsilon_{\text{soup}}(x_t, t, y)$ satisfies $\varepsilon$-NAF for some $\varepsilon$ which depends only on $D_1$ and $D_2$.
\end{proposition}
The above proposition suggests that, for different choices of $D_1$ and $D_2$ we can guarantee different subsets $C = \{c_i\} \subset D$ of NAF-protected samples. If all training samples require anti-memorization ($C = D$), then it suffices to pick $D_1$ and $D_2$ to be disjoint partitions of $D$. Conversely, in a mixed privacy setting \cite{golatkar2022mixed}, we have a \textit{public} set $D_\text{publ}$ set for which anti-memorization is not required, and a \textit{private} set $D_\text{priv}$ that requires anti-memorization. In this case, we can make better use of the data by picking $D_1 = D_\text{publ}$ to be a core safe set used to pre-train the model, and $D_2 = D_\text{publ} \cup D_\text{priv}$ to be the fine-tuning set. We explore both ideas for Diffusion models in Section \ref{sec:applications}.

\subsection{Computational efficiency}
Let $M$ denote the memory consumption (i.e. parameters) during inference, $T$ the time complexity of a forward pass. Compartmentalizing $n$ models has a worst case complexity of $\mathcal{O}(nM), \mathcal{O}(nT)$ for the memory and time respectively, while ensembling \cite{chen2022towards,zhou2022mixture,xue2024raphael} methods have $\mathcal{O}(nM), \mathcal{O}(T)$ complexity respectively. Diffusion Soup reduces the memory complexity by averaging the weights into a single model, resulting in an optimal $\mathcal{O}(M), \mathcal{O}(T)$ complexity. Souping reaps the benefits of multiple specialists and data sources without additional overhead.

\subsection{Unlearning} Given $n$ different data sources $D_f = \{D_i\}_i$, and their corresponding diffusion model weights $w_i$, Diffusion Soup provides a single set of weights $w_{\text{soup}} = \sum_i k_i w_i$. Just as adding new data sources to the soup is straightforward, removal is just as easy: should a data source $D_i$ decides to withdraw their data, we can simply update the souped weights as, $w_{\text{soup}} := \frac{w_{\text{soup}} - k_i w_i}{1-k_i}$. Should only a subset of data from $D_i$ be removed, we can simply reset the diffusion model $\epsilon_{w_i}$ to weights $w_c$, and fine-tune it with the remaining data in $D_i$. Souping thus enables efficient disgorgement of data.

\begin{table}[t!]
\caption{\textbf{Souping a collection of specialist models produces a high-quality generalist.} We consider data shards that allow the model to specialize in a range of image domains. Souping these specialist together outperforms all individual models. See Figure \ref{fig:splash} for a visualization of Diffusion Soup's performance. }
\label{table:mixture_10k_new}
\centering
\resizebox{0.85\columnwidth}{!}{%
\begin{tabularx}{\textwidth}{@{}>{\hsize=1.5\hsize}X >{\hsize=2.1\hsize}X *{3}{>{\hsize=0.46\hsize}C}@{}}

\toprule
&                                    
& \multicolumn{1}{c}{TIFA}     
& \multicolumn{1}{c}{IR}     
& \multicolumn{1}{c}{CLIP} \\

\midrule

\multirow{1}{*}{Base} 
& SD2.1 
& 85.5
& 0.30   
& 0.259 \\
\midrule

\multirow{9}{*}{SFT} 
& Animals (AN)      
& 86.0          
& 0.41
& 0.256 \\

& Body Parts (BP)   
& 85.6          
& 0.28
& 0.257 \\

& Electronics (EL) 
& 85.9           
& 0.42
& \textbf{0.260} \\

& Accessories (AC) 
& 85.3          
& 0.38
& 0.257 \\

& Clothes (CL) 
& 85.0            
& 0.35
& 0.256 \\

& Food (FO) 
& 85.7           
& 0.37
& 0.259 \\

& Hardlines (HA) 
& 85.6           
& 0.35
& \textbf{0.260} \\

& Products (PR) 
& 85.3            
& 0.37
& 0.258 \\

& Vehicles (VE) 
& 85.2            
& 0.33
& 0.257 \\

\midrule
\multirow{1}{*}{Combined}               
& All   
& 85.5           
& 0.34
& 0.257 \\

\midrule
\multirow{3}{*}{Souped}      
& Uniform
& 86.2          
& \textbf{0.45}
& \textbf{0.260}
\\
& Reverse Greedy
& 86.3
& 0.43
& \textbf{0.260}
\\
& Greedy
& \textbf{86.5}           
& 0.44
& 0.259
\\
\bottomrule
\end{tabularx}
}
\end{table}

\section{Experiments}
We empirically evaluate Diffusion Soup's ability to achieve training free continual learning and unlearning, provide anti-memorization guarantees, and blend disparate styles into unique hybrids.  Our experiments fall into five broad application domains: (1) specialist aggregation, (2) aesthetic enhancement, (3) unlearning (4) blending artistic styles, and (5) anti-memorization. We use Stable Diffusion 2.1 (SD2.1) as the pretrained model, and follow the finetuning procedure proposed in \cite{dai2023emu}. Our results are not unique to SD2.1, and broadly apply to any Diffusion Model. Details about hyperparameters are in the supplement.
\subsection{Datasets and Evaluation Metrics}
\subsubsection{Datasets} We use publicly-available datasets for our experiments. (1) \textit{Souping Specialists:} We generate category-specific datasets and finetune on them to obtain category specialists. These datasets are constructed by first using CLIP~\cite{radford2021learning} to retrieve the top 10K highest scoring images for each category, and then using the LAION aesthetic score~\cite{Schuhmann2022LAIONAesthetics} predictor to retain images with an aesthetic score above 6, leaving approximately a thousand images per category. (3) \textit{Style Mixing:} We use the Pokemon dataset \cite{pinkney2022pokemon} which contains images of Pokemon characters along with BLIP \cite{li2022blip} captions, and the FS-COCO dataset \cite{fscoco} to generate finetuned checkpoints. 
(4) \textit{Anti-Memorization and Unlearning:} We use the Pokemon dataset to evaluate anti-memorization and MSCOCO for unlearning. 

\begin{figure*}[t] 
\centering
\includegraphics[trim=0 175 300 0,clip,width=\linewidth]{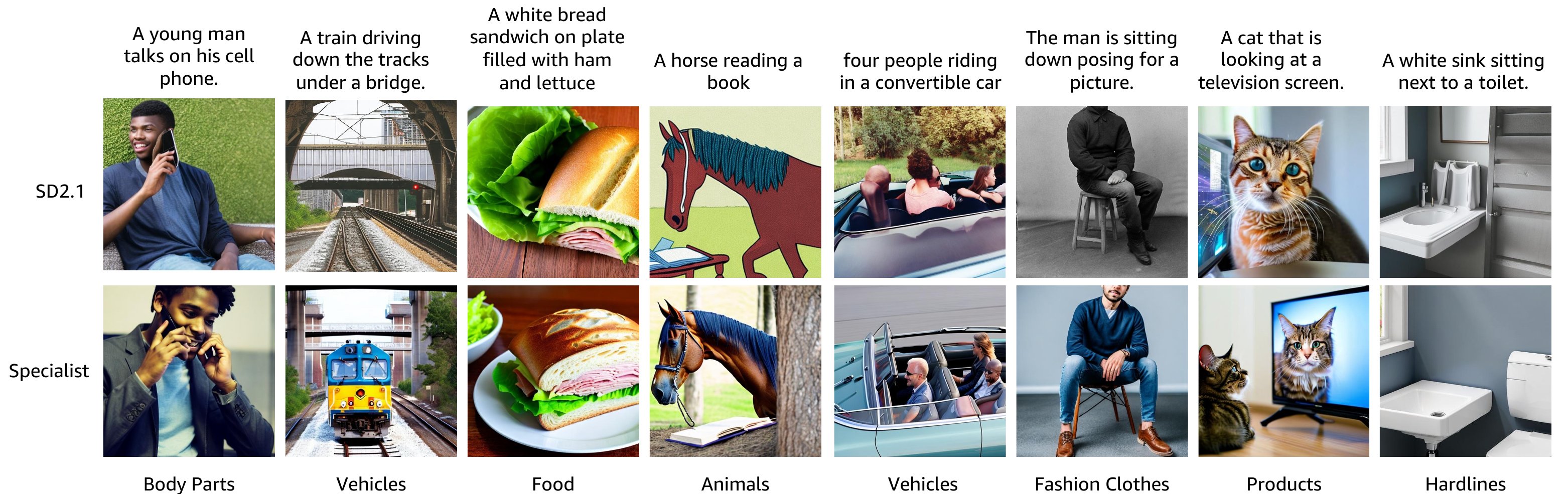}
\caption{\textbf{Finetuning Models on Category Specific Data Subsets Produces Specialists.} We visualize results from finetuning SD2.1 models on various data subsets shards by category. The finetuning process specializes the diffusion model to the subset's category: for example, the Body Parts dataset enhances the prominence of fingers, and the Fashion Clothes dataset enhances outfits. These models can be souped together to obtain generalized models that outperform all specialists (See Table \ref{table:mixture_10k_new}).}
\label{fig:specialists}
\end{figure*}

\subsubsection{Evaluation Metrics}
We consider three metrics commonly used in benchmarking the performance of diffusion models: (1) \textit{Text-to-Image Faithfulness evaluation with Question Answering}\cite{hu2023tifa} (TIFA Score) measures the faithfulness of a generated image to its text input via visual question answering (VQA), which we apply to 2k collected prompts from the MSCOCO Dataset.
(2) \textit{Image Reward} \cite{xu2024imagereward} (IR) is a scoring model that captures human preferences on image-text alignment, fidelity, and harmlessness. 
(3) \textit{CLIP Score} \cite{radford2021learning} (CLIP) captures text-image alignment via the cosine similarity between CLIP embeddings for the generated image and the text prompt. 

\subsection{Applications}\label{sec:applications}
\subsubsection{Continual Learning from Multiple Data Subsets}
The different data subsets that compartmentalization methods operate on are not chosen in practice but rather imposed by the changing nature of data. We therefore showcase the performance of Diffusion Soup on two broad groupings that data subsets often fall into: different \textit{domains} of data with different usage rights, or \textit{task-specific} data provided by different users. 

\paragraph{Soup of Specialists.} To represent different domains of data, we construct shards corresponding to diverse categories such as \emph{Animals}, \emph{Food}, \emph{Fashion Clothes}, etc and finetune SD2.1 to obtain domain specialists corresponding to each shard. \cref{fig:specialists} visualizes each specialist's ability to improve features corresponding to their domain over SD2.1, e.g. \emph{Animals} improve animal photorealism. We now compare various souping methods alongside these specialists and the paragon model trained on the union of all shards, and showcase our findings in Table \ref{table:mixture_10k_new}. We find that our finetuned specialists both outperform base SD2.1 in TIFA and IR, and often outperform the combined model paragon, confirming prior work \cite{golatkar2023training,balaji2022ediffi} demonstrating the value of training specialist models. A uniform soup of the specialist models outperforms all specialists and the paragon in TIFA, IR, and CLIP Score, indicating superior image quality and alignment compared to all base and individual specialist models. Our two greedy souping approaches further optimize TIFA and achieve the highest performing TIFA scores of 86.3 and 86.5. These image alignment optimizations come at a slight cost in image quality however, as indicated by the slight decrease in IR. Regardless, the approaches outperform all specialists and the combined model paragon in TIFA and IR, resulting in a Pareto optimal curve of souping options to choose from.
\begin{table}[t!]

\caption{\textbf{Diffusion Soup Outperforms Base and Combined Models in Aesthetic Enhancement}. We consider data shards that all have one task in common, aesthetic enhancement. Our souped results outperform all base models and the combined model in which a single model is trained on a union of the datasets.}
\label{table:scaling_emu_new2}

\centering

\label{tab:your-label}
\resizebox{0.8\columnwidth}{!}{%
\begin{tabularx}{\textwidth}{@{}l*{3}{C}|*{3}{C}|*{3}{C}@{}}


\toprule
& \multicolumn{3}{c}{Base \& SFT} 
& \multicolumn{3}{c}{Combined (Cumulative)} 
& \multicolumn{3}{c}{Souped (Cumulative)} \\
\cmidrule(lr){2-4} \cmidrule(lr){5-7} \cmidrule(lr){8-10}
& TIFA & IR & CLIP
& TIFA & IR & CLIP
& TIFA & IR & CLIP \\
\midrule
SD2.1 & 
85.5 & 0.30 & 0.259 &
- & - & - &
- & - & - \\
\midrule
AE1 & 


86.5 & 0.53 & \textbf{0.263} &
86.5 & 0.53 & \textbf{0.263} &
86.5 & 0.53 & \textbf{0.263} \\

AE2 & 

86.3 & 0.54 & 0.261 &
86.0 & 0.40 & 0.261 & 
\textbf{86.7} & \textbf{0.60} & \textbf{0.263} \\

AE3 & 

86.1 & 0.47 & 0.262 &
85.7 & 0.38 & 0.261 & 
\textbf{86.8} & \textbf{0.60} & \textbf{0.263} \\

AE4 & 

86.2 & 0.43 & 0.260 &
86.0 & 0.39 & 0.261 & 
\textbf{86.8} & \textbf{0.59} & \textbf{0.263}  \\

AE5 & 
85.9 & 0.45 & 0.262 & 
85.6 & 0.37 & 0.261 & 
\textbf{86.8} & \textbf{0.59} & \textbf{0.263} \\
\bottomrule
\end{tabularx}
}
\end{table}

We visualize the improvements of Diffusion Soup over the base model and the combined model paragon in \cref{fig:splash}, highlighting better image alignment, fewer artifacts, and improved aesthetics. Unlike other approaches such as ensembling or MoE methods, these improvements come without compromising on memory consumption and runtime, a significant advance for compartmentalization.
\paragraph{Enhancing Image Aesthetics.}\label{sec:aes-enhance} We next consider data shards all pertaining to a single task, aesthetic enhancement. We construct 5 highly aesthetic subsets of MSCOCO, AE1-5, ordered by decreasing aesthetic quality to simulate the diversity among data shards stemming from different providers. Once again, we compare various souping approaches to models trained on each shard, as well as a paragon trained on the union. To shed further light into the gap between the Diffusion Soup and the Paragon, we show \textit{cumulative} results for both as we take each aesthetic subset into consideration. 

Our results are highlighted in Table \ref{table:scaling_emu_new2}. We find that all finetuned models outperform base SD2.1 in TIFA and IR, but especially on IR, indicative of the aesthetic quality improvements that IR is particularly sensitive to. Once again, we find that a uniform Diffusion Soup outperforms base SD2.1, all finetuned models, and the combined paragon model on TIFA, IR and CLIP, respectively obtaining scores of 86.8, .59, and .263. The dominance of Diffusion Soup is also shown cumulatively, with the method outperforming the combined paragon model with every additional aesthetic dataset. As continual learning progresses over a growing number of dataset shards, Diffusion Soup continues to improve in performance and increase its lead on a combined model paragon. Curiously, we find that some individual models outperform the combined model paragon -- while these models are not specialists in the sense of the previous section, they appear to still specialize in the particular types of aesthetic enhancements present in each data shard, some of which better optimize for TIFA and IR than a combined dataset.

\subsubsection{Machine Unlearning}
\begin{figure*}[t] 
\centering
\includegraphics[trim=0 0 0 35, clip,width=\linewidth]{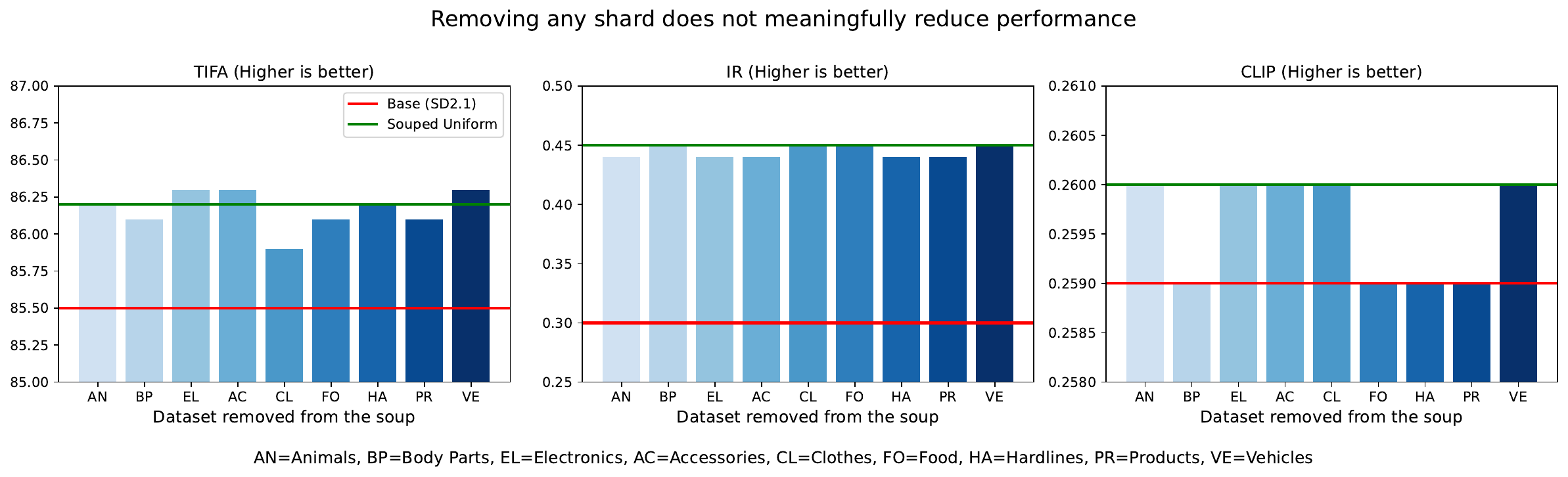}
\caption{\textbf{Removing Any Individual Data Shard Does Not Meaningfully Reduce Performance.} We soup specialists leaving one out at a time to demonstrate that no individual specialist significantly affects the quality of the generalist. Our results show that Diffusion Soup can be used for Machine Unlearning. From left-to-right, graphs show TIFA, Image Reward and CLIP Score. Performance of uniform soup model is in \emph{green}, and SD2.1 in \emph{red}.}
\label{fig:loosoe}
\end{figure*}

We next consider settings where data must be removed, or cannot be used anymore. For Diffusion Soup, just as adding data shards consisted simply of averaging models trained on those shards, removing shards can be trivially implemented through weighted subtraction. We demonstrate unlearning using our category datasets from which domain specialists are constructed, demonstrating that domain specialists can be removed from our model soup without meaningfully compromising the performance of the soup. Remarkably, we find that removing \textit{any} specialist reduces the performance of the soup by at most 1\% in IR. \cref{fig:loosoe} displays our full results, and shows that leaving any one shard out still produces a model soup that closely mirrors the performance of uniform soup in all metrics, and continues to outperform SD2.1 even in the worst case scenario, the removal of the \emph{Clothes} model. In some cases the removal of models improves the TIFA score, and this is the principle exploited by the Reverse Greedy Soup. However, in practice one cannot select the data that is removed, and we view our \cref{fig:loosoe} as supporting a uniform performance guarantee. The cumulative souping results of Table \ref{table:scaling_emu_new2} when read from bottom up serves as yet another demonstration of robust unlearning; indeed a model soup removing data shards AE4 and AE5 performs identically in TIFA score to the uniform soup over all shards, and further removal only results in a mild reduction in performance.

\subsubsection{Zero-Shot Style Mixing}
\begin{figure*}[t] 
\centering
\includegraphics[trim=0 175 200 0, clip,width=\linewidth]{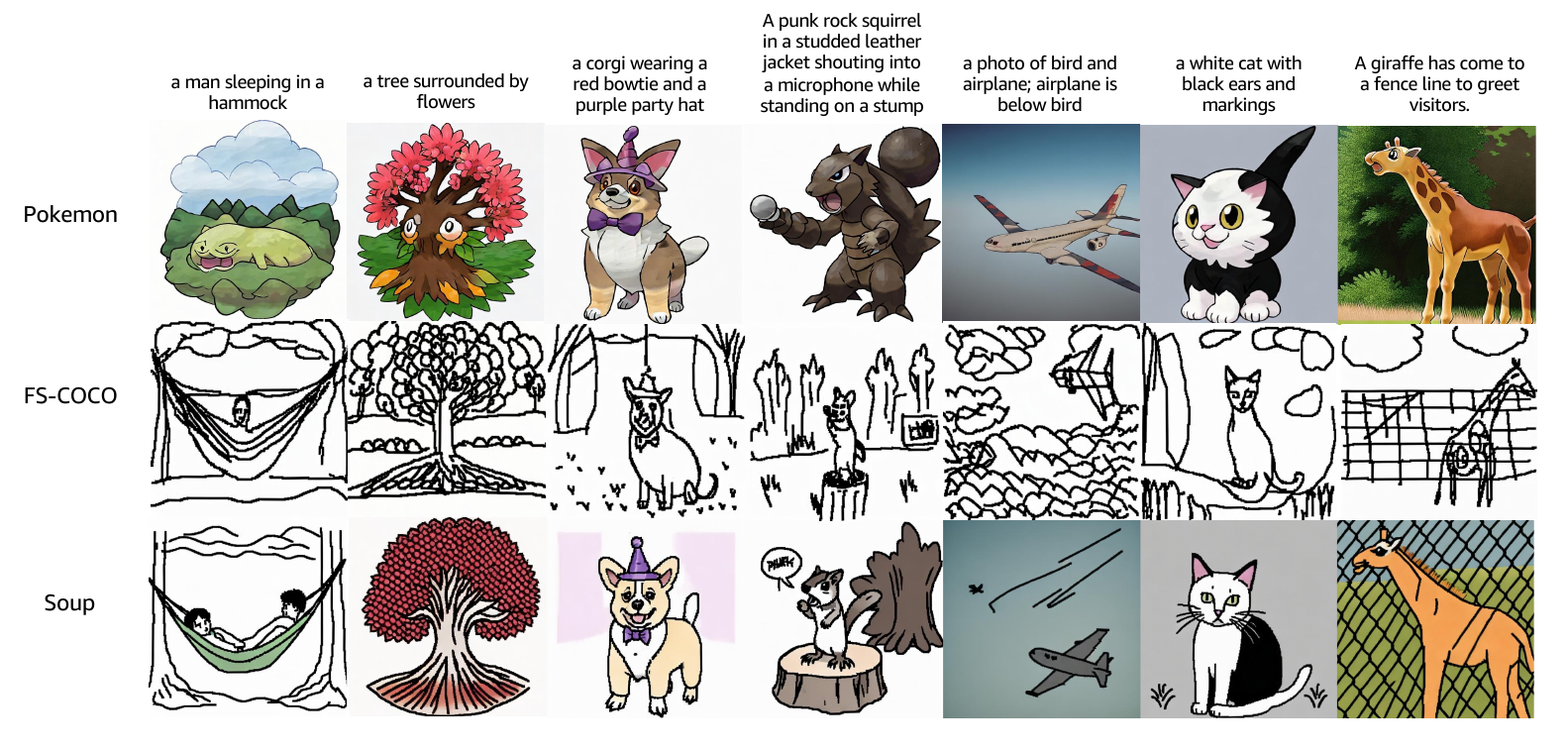}
\caption{\textbf{Diffusion Soup merges finetuned models to create hybrid styles.} We apply Diffusion Soup to models finetuned on Pokemon (\emph{Row 1}) and FS-COCO (\emph{Row 2}) to create a hybrid style (\emph{Row 3}). The results are zero-shot since we do not have examples of the hybrid style for training.}
\label{fig:style_mix_poke_fscoco}
\end{figure*}

We employ Diffusion Soup to merge the distinct styles of two separate datasets--- FS-COCO's sketch-like imagery and Pokemon illustrations---in a zero-shot manner, i.e. without the need for hybrid training samples. \cref{fig:style_mix_poke_fscoco} displays the results of this style fusion. The first row highlights the model tuned on the Pokemon dataset and the second row showcases the FS-COCO-tuned model which generates monochrome line-art sketches. The third row demonstrates that that souping these two models yields generation of a new hybrid style of line-art cartoon characters, without using any hybrid training examples. Our findings are the first to link model souping to Style Mixing, and stem from our novel application of souping to Diffusion Models, and more broadly generative models. We hypothesize that Style Mixing derives from Diffusion Soup sampling from the Geometric Mean of the constituent distributions, formalized in Proposition \ref{prop:geo_mean}, and therefore distinguishes our results from the outputs produced by training a model on the union of the two datasets. We leave further exploration of this hypothesis for future work. 

\begin{figure*}[t] 
\centering
\includegraphics[trim=0 340 0 10, clip,width=\linewidth]{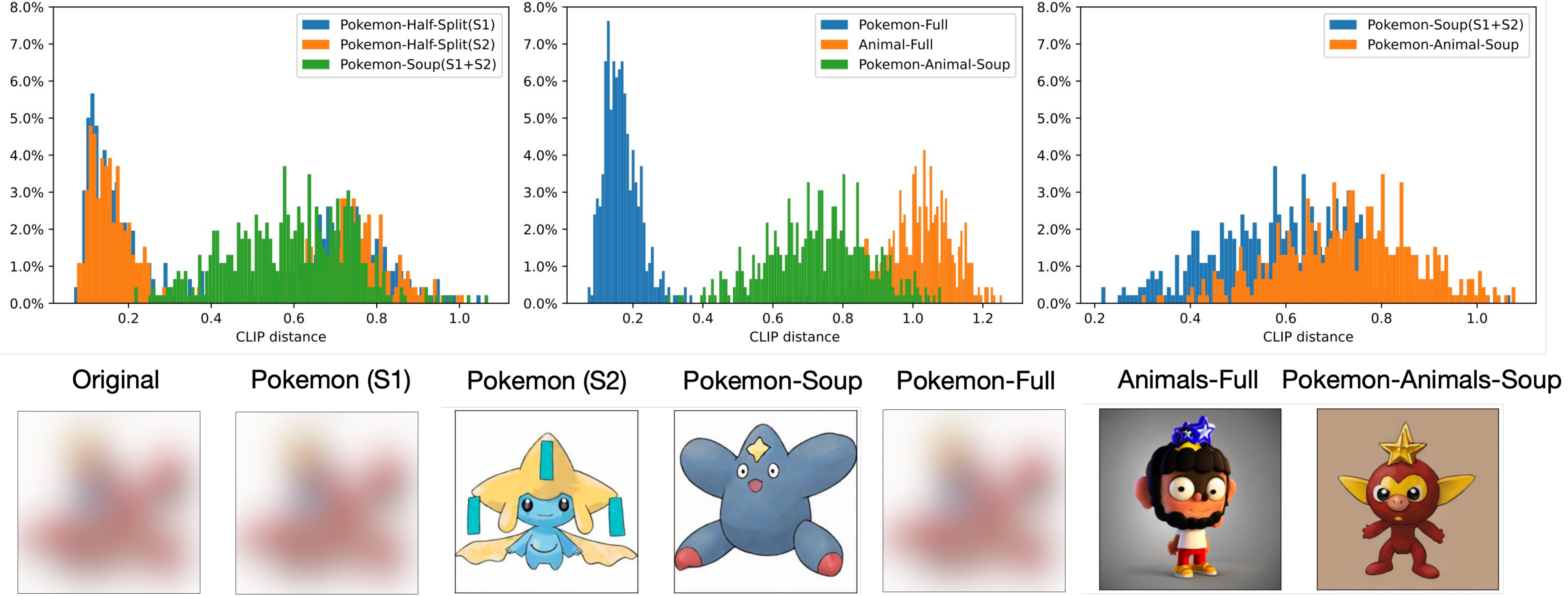}
\caption{\textbf{Diffusion Soup reduces dataset memorization.} \textbf{(Bottom)} Models (e.g. Pokemon-Full) trained on Pokemon data (Original) tend to reproduce some training samples at inference time. Note that we blur depictions of inputs in this subfigure. However, souping models (Pokemon-Soup) trained on disjoint subsets of Pokemon (S1 and S2) significantly reduces the memorization when compared to the Pokemon-Full. The same holds for cross-domain souping (Pokemon-\emph{Animals}-Soup), which soups the Pokemon model (Pokemon-Full) with a model trained on a safe dataset (\emph{Animals}-Full). \textbf{(Top)} We formalize this anecdotal result with a plot of the CLIP distance of various datasets to the Original Pokemon dataset (\emph{Top Left}). Larger distance or mass towards right implies lower memorization. Souping Pokemon subsets increases the CLIP distance (top middle) and therefore reduces reproduction of Pokemon. Souping Pokemon with \emph{Animals} reduces memorization further (\emph{Top Right}). \emph{Prompt used in figure: ``a cartoon character with a star on his head''}.}
\label{fig:soup_copy}
\end{figure*}
\subsubsection{Anti-Memorization}
We demonstrate that souping reduces memorization in Diffusion Models with two distinct scenarios, and display our results in \cref{fig:soup_copy}. In the first scenario, we randomly split the Pokemon dataset into two shards S1 and S2, and train separate models on each shard. Even though both models reproduce \textit{some} Pokemon, their soup surprisingly does not. We visualize an example in the bottom-middle of \cref{fig:soup_copy}, where Pokemon-Soup avoids reproducing samples from the Pokemon dataset while capturing its style. We formalize this finding at a distribution level using CLIP scores. Taking each prompt from the Original Pokemon Dataset, we measure the distance between the Original images and model generated images using CLIP scores. These values together chart a \textit{distance distribution} of model outputs to the original dataset, and we visualize such distributions in the top-left of \cref{fig:soup_copy}. There, the two models trained on shards (shown in \emph{blue} and \emph{orange}) each reproduce some samples from the original dataset, as evidenced by low CLIP score peaks of the distributions. The souped model's distance distribution (in \emph{green}) however is shifted considerably to the right, demonstrating that it generates images which are far from the Original data, i.e. avoids memorizing the dataset.

In the second scenario, we soup a model trained on the complete Pokemon dataset with a model trained on a completely disjoint domain---the \emph{Animals} model from Table \ref{table:mixture_10k_new}. We visualize an example in the bottom-left of \cref{fig:soup_copy}, where Pokemon-\emph{Animals}-Soup once again avoids memorization while capturing style. We plot distance distributions in the top-middle of \cref{fig:soup_copy}, where the Pokemon model (\emph{blue}) reproduces the Pokemon data, and the \emph{Animals} model (\emph{orange}) is unsurprisingly far away from it. The souped model's distribution (\emph{green}) again sits to the right of that of the Pokemon model, indicative of its reduced memorization. To compare the effect of the two souping approaches, we compare Pokemon-Soup and Pokemon-\emph{Animals} Soup's distance distributions in the top right of \cref{fig:soup_copy}, and find that the latter reduce memorization more than the former, but requires an additional dataset and corresponding model. These two methods are those both effective anti-memorization strategies, each with unique tradeoffs for a practitioner to navigate.

\subsubsection{Limitations} Diffusion Soup performs a variety of tasks with no additional inference cost, yet it still requires training $n$ model shards, and continual training on every new data shard. Also, averaging weights with a particular weighting of each model must be performed globally, and cannot be performed per sample without incurring the inference costs associated with ensembling. Although Adapters can help reduce both the training and reweighting burden, they come at the cost of reduced performance.

\section{Conclusion}
We present Diffusion Soup, a novel compartmentalization method in the image generation domain. Our method aggregates the benefits of many Text-to-Image models without the costly memory and inference time overhead of ensembling. Diffusion Soup outperforms constituent models, and a combined model paragon in specialist aggregation and aesthetic enhancement. Furthermore, Diffusion Soup enables two applications: the mixing of disparate artistic styles into novel hybrid styles and the ability to avoid memorizing images while capturing the underlying informational content. While our contribution reflects the first introduction of model souping to large scale generative AI, its potential to impact performance in other domains is untold. We leave as future work to explore souping in other domains such as language modeling, or visual question answering --- domains that can greatly magnify the impact of our work.

\clearpage  

%
%
\bibliographystyle{splncs04}
\bibliography{main}

\clearpage
\appendix
\section{Diffusion Soup: Model Merging for Text-to-Image Diffusion Models (Supplementary Material)}
\begin{figure*}[h] 
\centering
\includegraphics[width=0.49\linewidth]{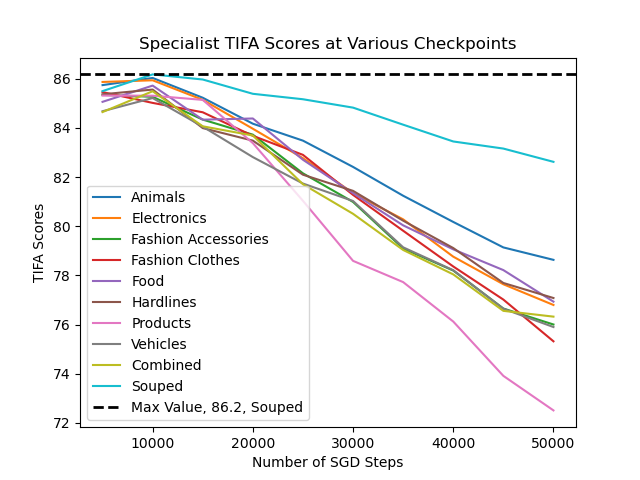}
\includegraphics[width=0.49\linewidth]{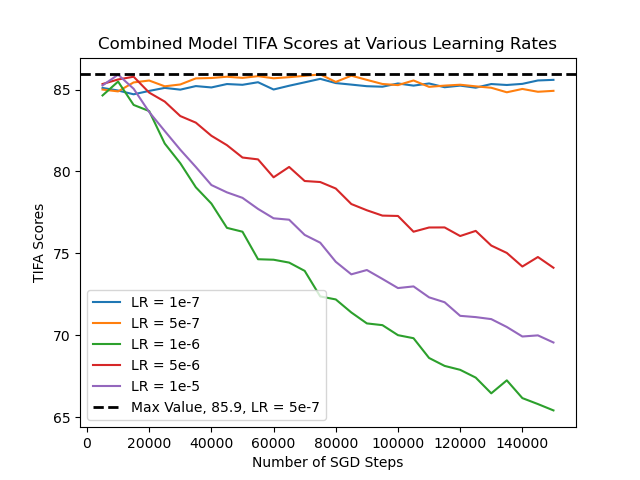}
\caption{\textbf{(Left) Diffusion Soup Outperforms All Models at All Checkpoints.} We show that optimizing over the number of SGD steps at LR = 1e-6 demonstrates that at all checkpoints, the souping model outperforms all individual models, as well as the combined models. The shallow dropoff in performance of the souped model relative to the steep dropoff of individual models at larger steps is indicative of strong robustness properties of Diffusion Soup. \textbf{(Right) Souping Outperforms the Paragon even when Optimizing the Latter for LR and Step Size Jointly.} We show that optimizing over Learning Rates and Step size jointly yields a better Combined Model Paragon (LR = 5e-7, Num Steps = 80k) than the original used in the main text (LR = 1e-6, Num Steps = 10k). However, Diffusion Soup still outperforms (TIFA 85.9 $\to$ 86.2) this new Combined Model Paragon checkpoint.}
\label{fig:sft_specialist_timestep}
\end{figure*}
The supplementary material is organized as follows. Appendix \ref{appsec:hyper} provides further details about hyperparameters used to conduct experiments in the paper, alongside ablations that justify our choice of hyperparameters.
Appendix \ref{appsec:ensemble} provides a comparison of Diffusion Soup to a computationally expensive Ensemble model. Appendix \ref{appsec:style_mix} provides additional examples of style mixing using compositions of famous artistic styles.
Appendix \ref{appsec:greedy} provides additional examples of greedy souping. Finally, Appendix \ref{appsec:proofs} gives
proofs for the main propositions of the paper.

\section{Choices of Hyperparameters and Ablations}\label{appsec:hyper}
The primary focus of our work is to showcase the benefits of Diffusion Soup atop a \textit{pre-existing} collection of finetuned checkpoints. Specifically, we aim to demonstrate that regardless of the ingredients provided, souping improves performance over any individual model, or the traditional approach of aggregating datasets to build one model (the Combined Paragon model). In this section, we show that regardless of the finetuned checkpoints, souping outperforms its constituent ingredients. We further optimize learning rates to find the strongest possible combined model paragon, to show the dominance of the souping approach even when the paragon is hyperparameter tuned at a disadvantage to souping.

To train our finetuned checkpoints for both the Specialist experiments and the Aesthetic Enhancement experiments, we use the AdamW optimizer with a Learning Rate of 1e-6. Our choice is mainly motivated by the Emu Paper \cite{dai2023emu} who recommend a low learning rate and finetuning for up to 15k steps to prevent model memorization and forgetting. We ablate the choice of the number of steps in \cref{fig:sft_specialist_timestep} (Left), and come to several conclusions. First, we show that as \cite{dai2023emu} suggest, the dropoff point for all finetuned models begins around 15k steps, and all models, including the Combined model peak at 10k steps. We thus use the 10k checkpoints for all of our individual and souping experiments in the main text. Moreover, we show that regardless of the number of steps chosen, the souped model always outperforms the constituent ingredients \textit{and} the combined model paragon, a powerful result that is indicative of the robustness of the souping approach. The horizontal black bar demonstrates that souping achieves the highest value overall, a TIFA score of $86.2$, the number we report in the main text for Uniform Souping.

We further ablate the choice of learning rate and step size for the Combined model Paragon in \cref{fig:sft_specialist_timestep} (Right), to ensure that souping is being compared to the strongest possible paragon baseline.
We find that a combination of an even lower learning rate (5e-7), and an extremely large step size of 80k steps outperforms (85.5 $\to$ 85.9) the combined model paragon at the learning rate of 1e-6 and step size of 10k that we use in the main text. However, our souped model at 1e-6 and step size of 10k still outperforms (85.9 $\to$ 86.2) this new Combined model checkpoint. 
We thus show that our souping methodology and findings are robust to hyperparameter tuning.

\begin{table*}[t]
\caption{\textbf{Diffusion Soup is Competitive even with a Computationally Expensive Ensembling Approach}. We revisit the setting of Table 2 in the main text, taking the first two aesthetic subsets AE1 and AE2, in order to compare Diffusion Soup with an ensembling approach. As observed previously, souping outperforms each constituent model and the combined paragon. We further show that the performance of our model is even competitive with a computationally costly Ensemble paragon, which computes the outputs of the denoising network using every model at every time step. The relative closeness of the two approaches, and the computational feasibilty of Diffusion Soup makes it a particularly attractive choice for practical users of compartmentalization.}

\label{table:scaling_emu_ensemble}
\centering
\resizebox{0.75\columnwidth}{!}{%
\begin{tabularx}{\textwidth}{@{}>{\hsize=0.75\hsize}X >{\hsize=0.75\hsize}X *{3}{>{\hsize=0.36\hsize}C}@{}}
\toprule
&
& TIFA
& IR
& CLIP \\
\midrule
\multirow{1}{*}{Base} 
& SD2.1
& 85.5 & 0.30 & 0.259 \\
\midrule

\multirow{2}{*}{SFT} 
& AE1
& 86.5
& 0.53
& \textbf{0.263} \\
                             
& AE2
& 86.3
& 0.54
& 0.261 \\
\midrule
\multirow{1}{*}{Combined (Paragon)}
& AE1 \& AE2
& 86.0
& 0.40
& 0.261 \\

\midrule
\multirow{1}{*}{Ensemble (Paragon)}
& AE1 \& AE2
& \textbf{87.3}
& \textbf{0.64}
& \textbf{0.263} \\
\midrule
\multirow{1}{*}{Souped (Ours)}
& AE1 \& AE2     
& 86.7
& 0.60
& \textbf{0.263} \\
\bottomrule
\end{tabularx}
}
\end{table*}

\section{Comparison to Ensembling}\label{appsec:ensemble}
In this section we compare Diffusion Soup against the \textit{ensembling} baselines \cite{golatkar2023training,du2023reduce,liu2022compositional}. These methods train different models on different subsets of data, and merge the flows---outputs of the denoising network, rather than the weights of the network as in souping---at inference. Precisely, ensembling approaches compute the output of the denoising network of each ingredient model at each time-step during backward diffusion and average them using weighted coefficients. We follow the approach in \cite{golatkar2023training} which shows that the weighted coefficients can be computed using a trained classifier.

Critically, ensembling approaches require \emph{each ingredient model to be loaded and run at inference time}. Therefore, they have a much larger memory footprint, and much greater inference overhead than Diffusion Soup, which generates a single model for use at inference time. To provide an anecdote illustrating the difference, in this paper we run all of our experiments on a single p4d.24xlarge EC2 instance in the AWS Cloud. Whereas our souping experiments often aggregate 5-9 model checkpoints together, we can only ensemble with 2 models without running into OOM errors on an A100 GPU.

We work within these limitations in \cref{table:scaling_emu_ensemble}, where we compare our method against the ensembling baselines. We take a pre-trained Stable-Diffusion model and fine-tune it in on two subsets of the MS-COCO dataset. Diffusion Soup is within $1\%$ ($86.7 \to 87.3$) TIFA and $93.75\%$ IR ($0.6 \to 0.64$) to ensembling. The small performance gap between the two approaches, and the computational feasibilty and scalability of Diffusion Soup makes souping a particularly attractive choice for practical users of compartmentalization.

\section{Additional Examples of Style Mixing}\label{appsec:style_mix}

\begin{figure*}[t] 
\centering
\includegraphics[page=1,trim=0 0 0 0,clip,width=\linewidth]{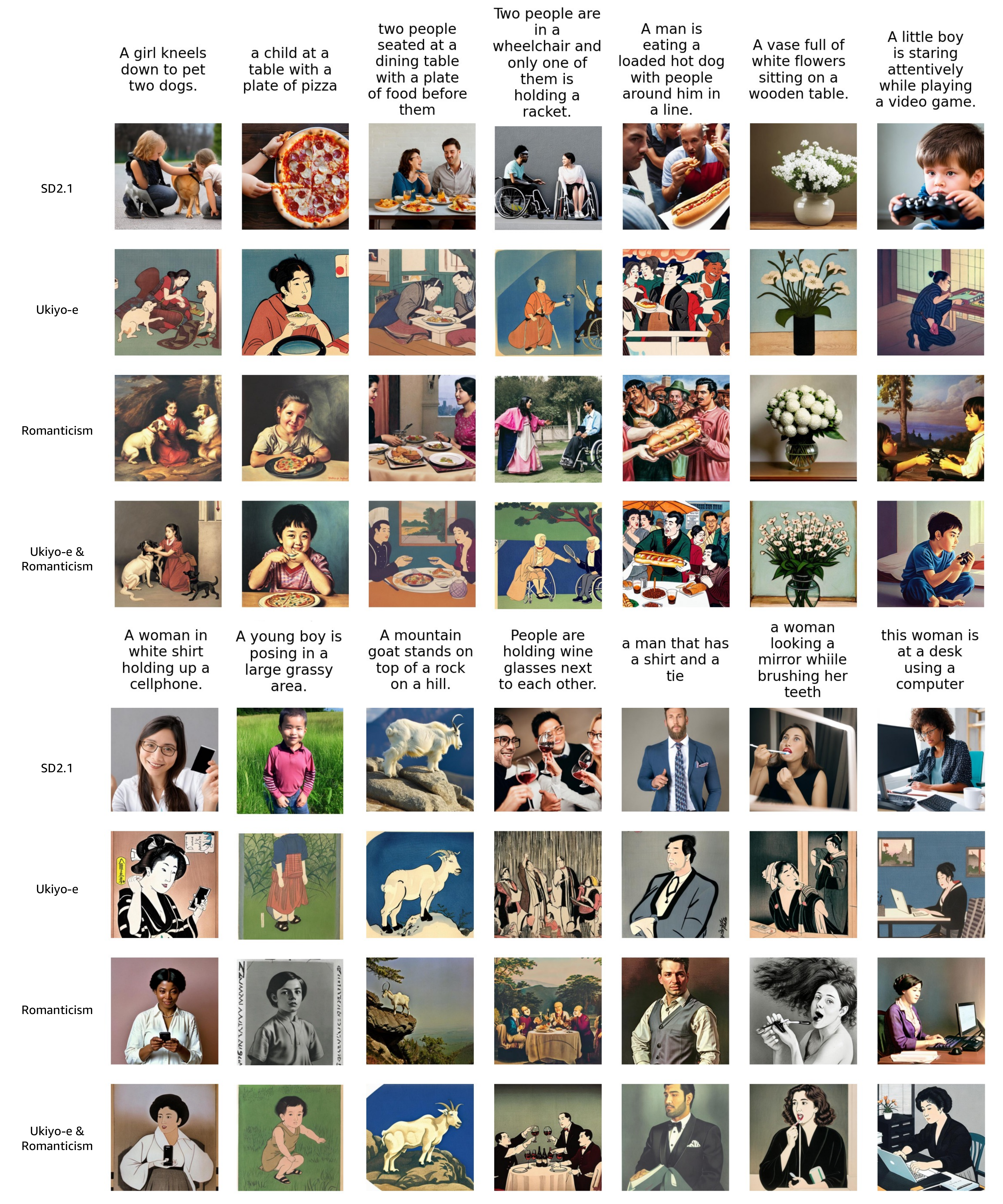}
\caption{\textbf{Diffusion Soup merges finetuned models to create hybrid artistic styles.} We apply Diffusion Soup to the SD2.1 model (\emph{Row 1}) finetuned on Ukiyo-e (\emph{Row 2}) and Romanticism (\emph{Row 3}) styles from WikiArt~\cite{artgan2018} to create a hybrid style (\emph{Row 4}). The results are zero-shot since we do not have examples of the hybrid style for training.}
\label{fig:stylemix_wikiart}
\end{figure*}

\begin{figure*}[t] 
\centering
\includegraphics[page=2,trim=0 25 0 0,clip,width=0.95\linewidth]{figures/stylemix/stylemix_extra_v2.pdf}
\includegraphics[page=1,trim=0 25 0 0,clip,width=0.95\linewidth]{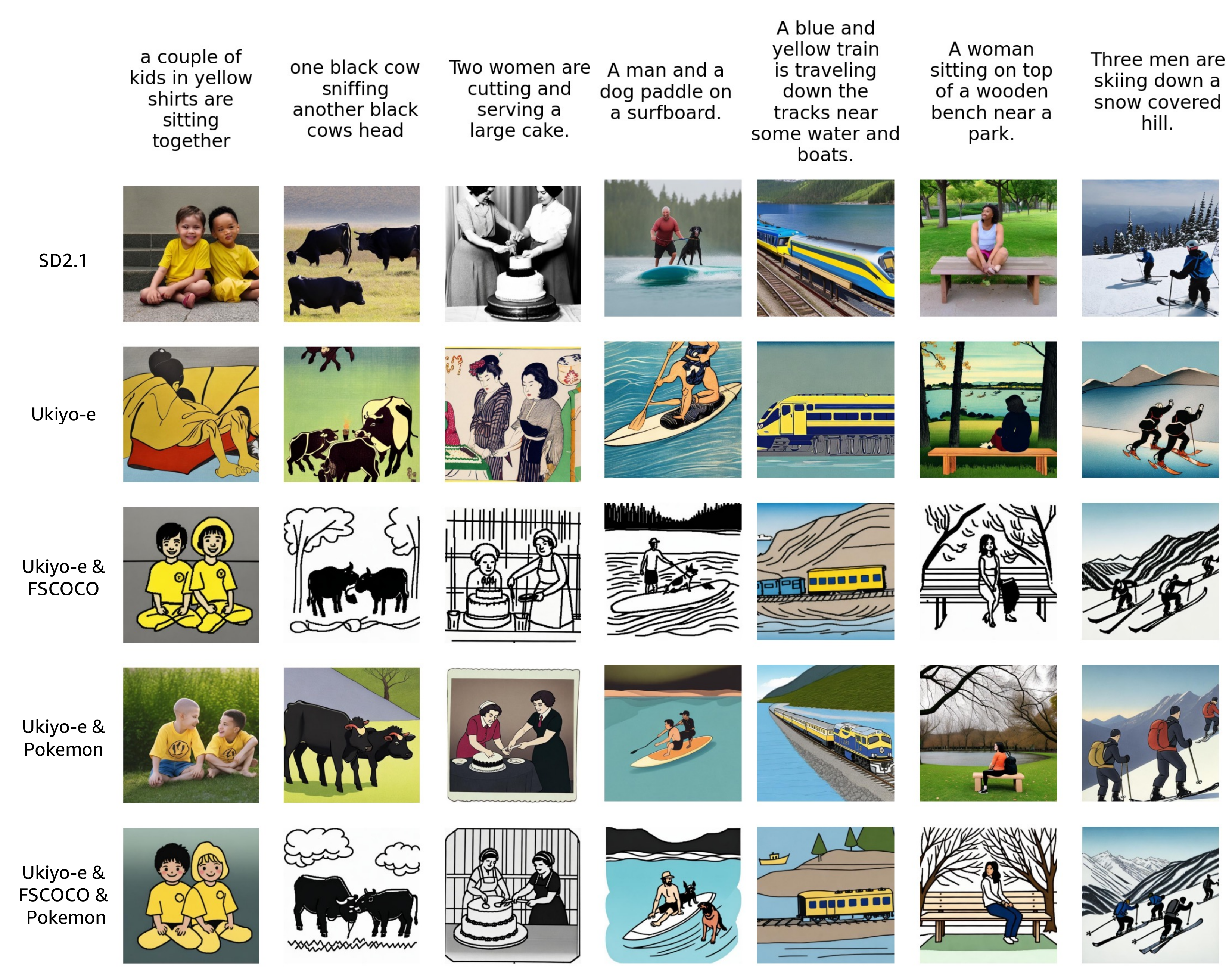}
\caption{\textbf{Diffusion Soup merges \emph{three} finetuned models to create hybrid styles.} We apply Diffusion Soup to the SD2.1 model (\emph{Row 1}) finetuned on Ukiyo-e (\emph{Row 2}) to create various hybrid styles. Ukiyo-e is souped with FSCOCO (\emph{Row 3}), Pokemon (\emph{Row 4}) and both models (\emph{Row 5}).}
\label{fig:stylemix_wikiart_pokemon}
\end{figure*}

We present additional zero-shot style mixing results using models fine-tuned on \emph{Ukiyo-e} and \emph{Romanticism} images from the WikiArt dataset~\cite{artgan2018}. \cref{fig:stylemix_wikiart} demonstrates souping Ukiyo-e and Romanticism styles. \cref{fig:stylemix_wikiart_pokemon} demonstrates blending the Ukiyo-e style with Pokemon~\cite{pinkney2022pokemon} and FSCOCO \cite{fscoco} styles. The final row of this figure demonstrates souping \emph{all three} styles.

\begin{table}[t]

\caption{\textbf{Greedy Souping Methods Outperform all Approaches in Aesthetic Enhancement}. We once again consider the setting of the main paper's Table 2, with data shards that all have one task in common, aesthetic enhancement. This time, we add our Greedy Souping approaches to Table 2 (R. refers to Reverse), showing that they outperform all other methods in Aesthetic Enhancement. These findings replicate the strong performance of Greedy Approaches found in aggregating specialists in the main text's Table 1.}
\label{table:scaling_emu_greedy}

\centering

\label{tab:your-label}
\resizebox{0.8\columnwidth}{!}{%
\begin{tabularx}{\textwidth}{@{}l*{3}{C}|*{3}{C}|*{3}{C}@{}}


\toprule
& \multicolumn{3}{c}{Base \& SFT} 
& \multicolumn{3}{c}{Combined (Cumulative)} 
& \multicolumn{3}{c}{Souped (Cumulative)} \\
\cmidrule(lr){2-4} \cmidrule(lr){5-7} \cmidrule(lr){8-10}
& TIFA & IR & CLIP
& TIFA & IR & CLIP
& TIFA & IR & CLIP \\
\midrule
SD2.1 & 
85.5 & 0.30 & 0.259 &
- & - & - &
- & - & - \\
\midrule
AE1 & 


86.5 & 0.53 & \textbf{0.263} &
86.5 & 0.53 & \textbf{0.263} &
86.5 & 0.53 & \textbf{0.263} \\

AE2 & 

86.3 & 0.54 & 0.261 &
86.0 & 0.40 & 0.261 & 
{86.7} & \textbf{0.60} & \textbf{0.263} \\

AE3 & 

86.1 & 0.47 & 0.262 &
85.7 & 0.38 & 0.261 & 
{86.8} & \textbf{0.60} & \textbf{0.263} \\

AE4 & 

86.2 & 0.43 & 0.260 &
86.0 & 0.39 & 0.261 & 
{86.8} & {0.59} & \textbf{0.263}  \\

AE5 & 
85.9 & 0.45 & 0.262 & 
85.6 & 0.37 & 0.261 & 
{86.8} & {0.59} & \textbf{0.263} \\
\bottomrule

Greedy Soup 
& - & - & - 
& - & - & - 
& \textbf{87.0} & \textbf{0.60} & \textbf{0.263} \\
R. Greedy Soup 
& - & - & - 
& - & - & - 
&86.9  & 0.59 & \textbf{0.263} \\
\bottomrule
\end{tabularx}
}
\end{table}

\section{Greedy souping for \emph{Enhancing Image Aesthetics}}\label{appsec:greedy}

We extend the experimental section from~\cref{sec:aes-enhance} and~\cref{table:scaling_emu_new2} with Greedy Souping results (see~\cref{sec:greedysoup}). The original cumulative findings showed that Uniform Soup outperforms its ingredients in a cumulative sense, i.e. as the number of checkpoints grow from AE1 all the way to AE1-AE5. Our greedy approaches strategically select a \text{subset} of these checkpoints to show that performance can be further optimized in TIFA score. Identical to the setting with Specialists, Greedy Souping outperforms all other souping approaches, and both Greedy Souping and Reverse Greedy Souping are the two highest performing approaches. Our strong performance demonstrates that Greedy Souping offers benefits when merging models trained on a common task (in this case aesthetic enhancements) as well as aggregating specialists (see~\cref{table:mixture_10k_new}).

\section{Proofs}\label{appsec:proofs}
We restate each proposition for convenience prior to providing their proof.
\setcounter{proposition}{0}
\subsection{Proposition 1}
\begin{proposition}(Geometric Mean)\label{prop:geo_mean}
Let $\nabla_{x_t} \log p^{(i)}(x_t)$ be the marginal score (in \cref{eq:backward-diffusion}) where  $x_t = \gamma_t x_0 + \sigma_t \epsilon$, with $x_0 \sim p^{(i)}(x_0)$ and $\epsilon \sim \mathcal{N}(0, I)$. Let $\epsilon_{w_i}(x_t, t, y)$ be a neural network (with sufficient capacity) trained to match $\nabla_{x_t} \log p^{(i)}(x_t)$. Under certain conditions on the sampling procedure \cref{eq:backward-diffusion}, using $(1/n)\sum_{i=1}^n\nabla_{x_t} \log p^{(i)}(x_t)$ generates samples from $Z^{-1}\big(\Pi_i p^{(i)}(x_0)\big)^{1/n}$. Furthermore, using \cref{eq:linearization}, $\epsilon_{\text{soup}}$ with $k_i = 1/n$ generates samples from $Z^{-1}\big(\Pi_i p^{(i)}(x_0)\big)^{1/n}$.
\end{proposition}
\textbf{Proof}
Let $p_{\text{geometric}} = Z^{-1}\big(\Pi_i p^{(i)}(x_0)\big)^{1/n}$ be the geometric mean of the initial set of distributions provided to us. The marginal distribution at time $t$ corresponding to the geometric mean is given by $p(x_t) = \int Z^{-1}\big(\Pi_i p^{(i)}(x_0)\big)^{1/n} p(x_t|x_0) dx_0$. Sampling from this distribution requires access to the score of the marginal, which is usually difficult to obtain. This score is given by:
\[\nabla_{x_t} \log p(x_t) = \nabla_{x_t} \log \int Z^{-1}\big(\Pi_i p^{(i)}(x_0)\big)^{1/n} p(x_t|x_0) dx_0 \]
where the integral inside the logarithm is difficult to estimate. However, we can show that we can construct a simple score function which is easy to estimate and can generate samples from $p_{\text{geometric}}$:
\[\nabla_{x_t} \log p'(x_t) = (1/n)\sum_i \nabla_{x_t} \log \int p(x_t|x_0) p^{(i)}(x_0) dx_0 \]
$\nabla_{x_t} \log p'(x_t)$ can be used for sampling because of the fact that it generates a sequence of distributions where its equal to $\nabla_{x_t} \log p(x_t)$ at $t=0$, and $\mathcal{N}(0,I)$ at $t=T$. Thus using $\nabla_{x_t} \log p'(x_t)$ we can create a sequence of distributions which generates a sample from $p_{\text{geometric}}$ as $t \rightarrow 0$ during backward diffusion using \cref{eq:backward-diffusion}. Note that this sequence of distributions is different from the sequence of distributions generated by the forward diffusion process. Convergence of langevin based MCMC (using \cref{eq:backward-diffusion} for sampling) is a very well studied problem \cite{yang2022convergence,cheng2018convergence,vempala2019rapid,neal2001annealed}, showing that using \cref{eq:backward-diffusion} indeed converges to stationary distribution given by $p_{\text{geometric}}$ in this case. Furthermore,\cite{yang2022convergence} in Theorem 1 shows that sampling using Langevin dynamics (\cref{eq:backward-diffusion}) with a sufficiently powerful score estimator (in our case a diffusion model) generates samples from the initial distribution, $p_{\text{geometric}}$. This proves the first part of the result. The second part of the proof is a mere application of \cref{eq:linearization} and \cref{eq:model-soup} i.e. we replace each $\nabla_{x_t} \log p^{(i)}(x_t)$ with $-\epsilon_{w_i}(x_t, t, y)$. More precisely,
\begin{equation}
\nabla_{x_t} \log p(x_t) = (1/n) \sum_i -\epsilon_{w_i}(x_t, t, y) = -\epsilon_{\text{soup}} 
\end{equation}

\subsection{Proposition 2}
\begin{proposition}(Arithmetic Mean)
Let $\nabla_{x_t} \log p^{(i)}(x_t)$ be the marginal score (in \cref{eq:backward-diffusion}) where  $x_t = \gamma_t x_0 + \sigma_t \epsilon$, with $x_0 \sim p^{(i)}(x_0)$ and $\epsilon \sim \mathcal{N}(0, I)$. Let $\epsilon_{w_i}(x_t, t, y)$ be a neural network (with sufficient capacity) trained to match $\nabla_{x_t} \log p^{(i)}(x_t)$. Sampling using $\sum_{i=1}^n \lambda_i \frac{p^{(i)}(x_t)}{p(x_t)}\nabla_{x_t} \log p^{(i)}(x_t)$ with \cref{eq:backward-diffusion} generates samples from $\sum_i \lambda_i p^{(i)}(x_0)$, where $p(x_t) = \sum_i \lambda_i p^{(i)}(x_t)$, $\sum_i \lambda_i = 1$. Furthermore, using \cref{eq:linearization}, $\epsilon_{\text{soup}}$ with $k_i = \lambda_i \frac{p^{(i)}(x_t)}{p(x_t)}$ generates samples from $\sum_i \lambda_i p^{(i)}(x_0)$.
\end{proposition}
\textbf{Proof}
Let $\nabla_{x_t} \log p(x_t)$ be the score the mixture of distributions $p(x_t) = \sum_i \lambda_i p^{(i)}(x_t)$. We show that the score of the mixture is a convex combination of the individual scores:
\begin{align}
\nabla_{x_t} \log p(x_t) &= \nabla_{x_t} \log \sum_i \lambda_i p^{(i)}(x_t) \nonumber \\
&=\dfrac{1}{\sum_i \lambda_i p^{(i)}(x_t) } \sum_i \lambda_i \nabla_{x_t} p^{(i)}(x_t) \nonumber \\
&=\dfrac{1}{\sum_i \lambda_i p^{(i)}(x_t) } \sum_i \lambda_i \dfrac{p^{(i)}(x_t)}{p^{(i)}(x_t)} \nabla_{x_t} p^{(i)}(x_t) \nonumber \\
&=\dfrac{1}{\sum_i \lambda_i p^{(i)}(x_t) } \sum_i \lambda_i p^{(i)}(x_t) \nabla_{x_t} \log p^{(i)}(x_t) \nonumber \\
&=\sum_i \dfrac{\lambda_i p^{(i)}(x_t)}{\sum_i \lambda_i p^{(i)}(x_t) } \nabla_{x_t} \log p^{(i)}(x_t) \nonumber \\
\implies \nabla_{x_t} \log p(x_t) &= \sum_i \dfrac{\lambda_i p^{(i)}(x_t)}{\sum_i \lambda_i p^{(i)}(x_t) } \nabla_{x_t} \log p^{(i)}(x_t) \nonumber
\end{align}
Hence, we show the first part of the proof. The second part of the proof is a mere application of \cref{eq:linearization} and \cref{eq:model-soup} i.e. we replace each $\nabla_{x_t} \log p^{(i)}(x_t)$ with $-\epsilon_{w_i}(x_t, t, y)$:
\begin{align}
\nabla_{x_t} \log p(x_t) &= \sum_i \dfrac{\lambda_i p^{(i)}(x_t)}{\sum_i \lambda_i p^{(i)}(x_t) } (-\epsilon_{w_i}(x_t, t, y)) \nonumber \\
&= -\epsilon_{\text{soup}} \nonumber \\
\end{align}
where we use \cref{eq:model-soup} to show the previous equality with $k_i=\dfrac{\lambda_i p^{(i)}(x_t)}{\sum_i \lambda_i p^{(i)}(x_t)}$.

\subsection{Proposition 3}
\begin{proposition}($\varepsilon$-NAF for Diff.\ Soup)
Let $\epsilon_{\text{soup}}(x_t, t, y) \triangleq \epsilon_{\frac{1}{2}(w_1+w_2)}(x_t, t, y)$ be the soup model obtained by training $w_i$ on $D_i$, such that $D = D_1 \cup D_2$, and $D_1 \cap D_2 = \varnothing$. Then, under certain conditions sampling, using $\epsilon_{\text{soup}}(x_t, t, y)$ satisfies $\varepsilon$-NAF for some $\varepsilon$ which depends only on $D_1$ and $D_2$.
\end{proposition}
\textbf{Proof}
\cite{vyas2023provable} shows that given two generative models $p_1(x|y), p_2(x|y)$, sampling from $\sqrt{p_1(x|y)p_2(x|y)}/Z$ produces data which is NAF with respect to some protected samples in the training data. We need to show that souping generates samples from $\sqrt{p_1(x|y)p_2(x|y)}/Z$. Note that this directly follows from \cref{prop:geometric} for the case $n=2$.

\end{document}